\documentclass[conference]{IEEEtran}
\IEEEoverridecommandlockouts

\usepackage{cite}
\usepackage{amsmath,amssymb,amsfonts}
\usepackage{graphicx}
\usepackage{textcomp}
\usepackage{xcolor}
\usepackage{amsmath,amsfonts}

\usepackage{array}
\usepackage{textcomp}
\usepackage{stfloats}
\usepackage{url}
\usepackage{verbatim}
\usepackage{graphicx}
\usepackage{cite}
\usepackage{booktabs}
\usepackage{subfigure}
\usepackage{booktabs}
\usepackage[ruled]{algorithm2e}
\def\BibTeX{{\rm B\kern-.05em{\sc i\kern-.025em b}\kern-.08em
    T\kern-.1667em\lower.7ex\hbox{E}\kern-.125emX}}
\begin{document}

\title{FedCCA: Client-Centric Adaptation against Data Heterogeneity in Federated Learning on IoT Devices
\thanks{This work is supported in part by Hong Kong RGC Theme-based Research Scheme (No.: T43-
513/23-N); in part by Hong Kong RGC Collaborative
Research Fund (No.: C5032-23GF); and in part by the Research Institute for Artificial Intelligence of Things, The Hong Kong Polytechnic University.}
}

\author{
\IEEEauthorblockN{Kaile Wang$^{\dag}$, Jiannong Cao$^{\dag}$, Yu Yang$^{\ddag}$, Xiaoyin Li$^{\dag}$, Yinfeng Cao$^{\dag}$}
\IEEEauthorblockA{\textit{Department of Computing, The Hong Kong Polytechnic University, Hong Kong, China$^{\dag}$} \\
\textit{Centre for Learning, Teaching, and Technology, The Education University of Hong Kong, Hong Kong, China$^{\ddag}$}\\
}
}

\maketitle

\begin{abstract}

With the rapid development of the Internet of Things (IoT), AI model training on private data such as human sensing data is highly desired. Federated learning (FL) has emerged as a privacy-preserving distributed training framework for this purpuse. However, the data heterogeneity issue among IoT devices can significantly degrade the model performance and convergence speed in FL. Existing approaches limit in fixed client selection and aggregation on cloud server, making the privacy-preserving extraction of client-specific information during local training challenging. To this end, we propose Client-Centric Adaptation federated learning (FedCCA), an algorithm that optimally utilizes client-specific knowledge to learn a unique model for each client through selective adaptation, aiming to alleviate the influence of data heterogeneity. Specifically, FedCCA employs dynamic client selection and adaptive aggregation based on the additional client-specific encoder. To enhance multi-source knowledge transfer, we adopt an attention-based global aggregation strategy. We conducted extensive experiments on diverse datasets to assess the efficacy of FedCCA. The experimental results demonstrate that our approach exhibits a substantial performance advantage over competing baselines in addressing this specific problem.

\end{abstract}


\section{Introduction}
The rapid development of the Internet of Things (IoT) has revolutionized the way devices interact, communicate, and process data in a wide range of applications, from smart homes to industrial automation. As IoT devices continuously generate and exchange vast amounts of sensitive information, the demand for robust privacy protection mechanisms has become increasingly critical. Ensuring the confidentiality and integrity of user data is paramount, especially in scenarios where data is distributed across numerous devices with varying capabilities and security requirements. 
Federated learning (FL) has emerged as a promising solution to address privacy concerns in distributed machine learning environments \cite{9415623, 9460016}. By enabling collaborative model training without the need to centralize raw data, FL preserves data privacy and reduces the risk of data leakage. In this framework, individual devices, or clients, perform local training on their private data and only share model updates with a central server, which aggregates these updates to improve the global model.

\begin{figure}[t] 
  \centering
  \includegraphics[width=0.5\linewidth]{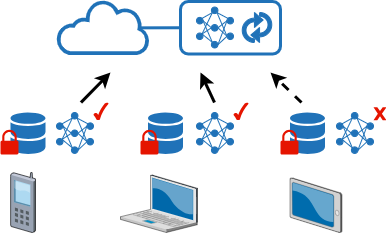}
  \caption{Illustrative examples of heterogeneous scenarios in federated learning on IoT devices.}
  \label{fig:illustration} 
\end{figure}

Despite the progress distributed machine learning algorithms have achieved, one of the eminent challenges federated learning (FL) faces is data heterogeneity when deployed in real-world IoT settings, as shown in Fig. \ref{fig:illustration}. In the real world, clients data in most cases are not \textit{independent and identically distributed (non-IID}) e.g., data collected by different sensors, wearable devices, and hospitals, or images taken in different scenarios and styles, leading to domain shift between clients \cite{10.5555/1462129}. 
As federated learning is proposed as a promising privacy-preserving paradigm enabling participants to jointly learn a global model without data sharing \cite{Fedavg}, such non-IID setting violates the basic assumption of conventional federated learning algorithms and as a result may lead to degradation in model performance and failures in model convergence. 
Generally, the non-IID scenes in federated learning can be categorized into label shift and domain shift.
This heterogeneity often leads to degraded model performance and slower convergence rates, undermining the effectiveness of federated learning in practical applications.

While a myriad of methods have been proposed to handle the above issue \cite{Huang_2022_CVPR, Huang_Chu_Zhou_Wang_Liu_Pei_Zhang_2021, li2021fedbn}, existing approaches have been overlooking the importance of dynamic, client-centric strategies for model aggregation on the cloud server. In practical IoT environments, each client may possess unique data characteristics and requirements, making a static selection approach suboptimal. Furthermore, effectively extracting and utilizing client-specific information during local training is crucial for personalizing models and enhancing overall performance. However, achieving this personalization without compromising the privacy of client data remains a significant challenge. The need to balance the extraction of client-specific knowledge with privacy constraints highlights a critical gap in current federated learning methodologies, especially in the context of heterogeneous and privacy-sensitive IoT applications. 

Notably, the property of the selection of clients is not well exploited, as previous works \cite{pmlr-v151-jee-cho22a} mainly focus on system-level optimization strategy. Although several selection criteria have been investigated in recent literature \cite{8761315,MLSYS2020_38af8613}, they place importance either on eliminating inefficiency and redundancy or improving fairness and client contribution. Thus, ignoring the fact that in more realistic and large-scale federated setting, client selection is usually random and data distribution is more heterogeneous. The very objective of personalized federated learning prioritizes optimization for each client over system-level effectiveness. In addition, several approaches require an unrealistic setting where a centralized dataset on the server is regarded as source domain and the pre-trained source model is adapted to client data \cite{Yao_2022_WACV}, which rarely holds for real-world cases. 
The above insights encourage an overlooked need for selective adaptation from a client-centric perspective.

Motivated by the aforementioned discoveries, we propose a novel method called client-centric Domain Adaptation federated
learning (FedCCA) to address the non-IID issue in federated learning on IoT devices. FedCCA optimally utilize client-specific knowledge to learn an unique model for each client through selective adaptation, aiming to alleviate the influence of data heterogeneity. Specifically, FedCCA introduces an additional client-specific encoder for each device, which is designed to extract and encode personalized features from local data during training. Meanwhile, FedCCA employs dynamic client selection and adaptive aggregation based on an additional client-specific encoder, which enables the extraction of personalized features while preserving data privacy. By tailoring the client selection process to the characteristics of each client, our approach ensures that the selected clients better reflect the diverse data distributions present across the network. Furthermore, to enhance the transfer of knowledge from multiple sources, we adopt an attention-based global aggregation strategy. This mechanism allows the model to weigh the contributions of different clients according to their relevance and informativeness, thereby improving overall performance and convergence. Through these innovations, FedCCA provides a robust solution for federated learning in heterogeneous IoT environments, facilitating more accurate and personalized model training without compromising privacy.

In conclusion, our main contributions in this work are as follows:
\begin{itemize}
\item We tackle the challenging problem of data heterogeneity in federated learning on IoT devices and formulate a practical problem setting, multi-domain heterogeneous federated adaptation. To address the above issue, we explore the significance of client selection and global aggregation strategies.
\item We propose FedCCA, a novel client-centric Adaptation federated learning algorithm specifically designed to address the issue. FedCCA incorporates dynamic client selection and attention-based aggregation strategies to leverage the most informative and relevant client updates, leading to improved model performance and convergence in federated learning.
\item We conduct extensive experiments on real-world datasets. Compared to baseline methods, we improve local model performance on all tasks, demonstrating significant practical improvements on the extensive experiments.
\end{itemize}

\section{RELATED WORK}

\subsection{Federated Learning on Non-IID Data}
The pioneering federated learning method FedAvg \cite{Fedavg} enables multiple clients to collaboratively learn a global model without sharing private data. 
However, the data heterogeneity across clients imposes challenges for the system to obtain an unbiased estimate of the full gradient. Extensive efforts \cite{li2021fedbn} have been made to address the non-IID issue. Existing works tackle the challenge through adding a proximal term to \cite{Fedavg} for partial information aggregation\cite{MLSYS2020_38af8613}. To mitigate the impact of data heterogeneity, personalized federated learning has emerged these years training models for distinct data distribution from each client. In this line of work, some researchers employ personalization layer with unique local heads for each client \cite{collins2021exploiting}, while other alternative approaches including fine-tuning \cite{wang2019federated}, multi-task learning and meta-learning \cite{NEURIPS2019_f4aa0dd9} also enjoy delightful success.
More recently, aiming at alleviating domain shift between client data, the combination of representation learning and federated transfer learning has gained increasing attention \cite{zhang2024eliminating,crawshaw2024federated}. Meanwhile, a new setting, namely domain-mixed federated learning, assumes client dataset with multiple domains is discussed \cite{zhong2023feddar}. The idea of learning a simple representation of the data is prompted using restrictions of the amount of information the representation can contain with two common regularizers \cite{nguyen2022fedsr}. A federated learning based cross-domain recommendation system is proposed to design a personal module and a transfer module to adapt to the extremely heterogeneous data on the participating devices \cite{10.1145/3511808.3557320}. As feature extraction being a crucial signature under this concept, some works propose to learn a shared global model in a layer-wise manner by matching and averaging hidden elements \cite{Wang2020Federated}. While \cite{10.1145/3511808.3557378} propose to achieve personalized federated optimization with meta
critic to capture robust and generalizable domain-invariant knowledge across clients.

\subsection{Client Selection for Federated Learning}
In most of the federated learning frameworks, not all clients are selected to join the aggregation in each communication round due to limited communication capacity. \cite{8761315} proposed a Partial participation strategy and stress that the number of updated clients in each round would affect the overall performance. 
As previous works mostly assumed unbiased client participation and random selection or in proportion of their data sizes, \cite{cho2021client} presented the first convergence analysis of federated optimization for biased client selection strategies, and quantified how the selection bias affects convergence speed. \cite{9766408} formulated the client selection problem under joint consideration of effective participation and fairness. 
In a more recent work, \cite{10197174} systematically presented recent advances in the emerging field of Federated Learning client selection and its challenges and research opportunities for the client heterogeneity problem. 
While numerous studies have been carried out, the discussion mostly concentrated on the server level. None of these works treat a single client as the focus of the selection strategy. A client-based selection strategy has not been thoroughly discussed.

\begin{figure*} 
  \centering
  \includegraphics[width=0.8\linewidth]{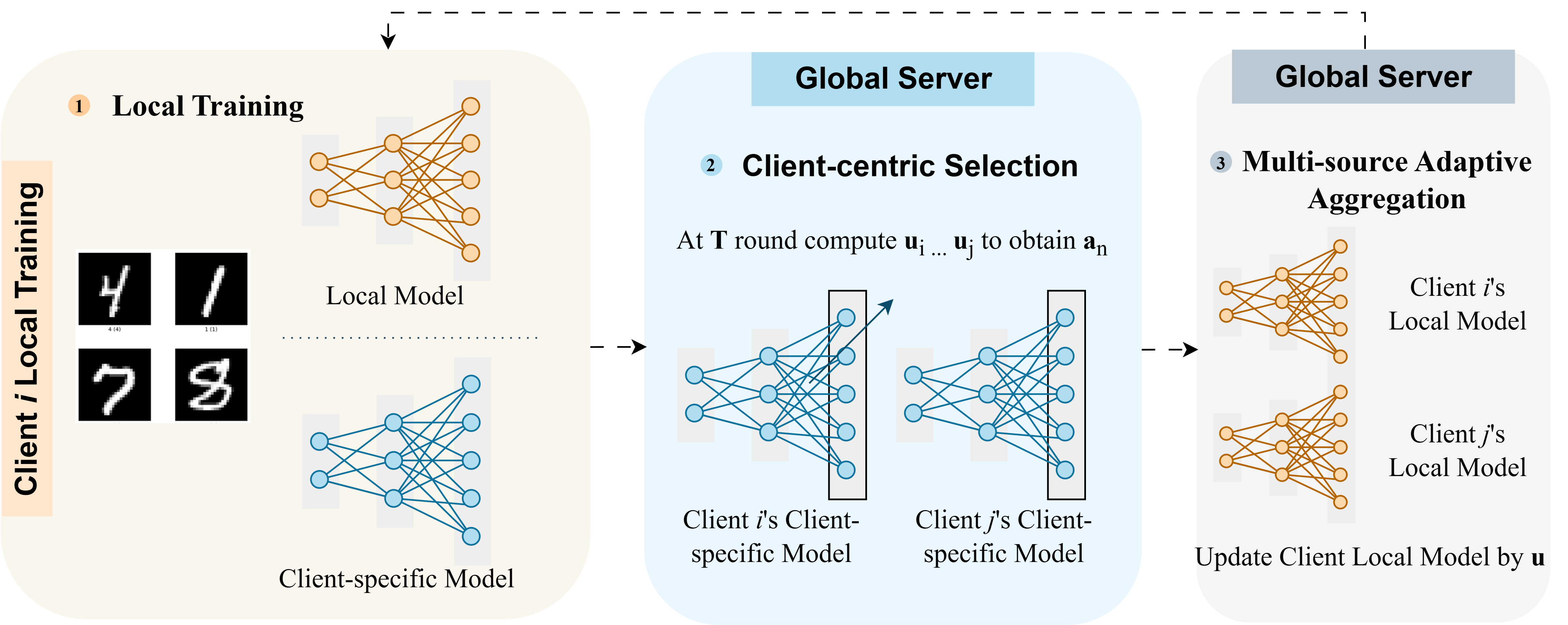}
  \caption{FedCCA algorithm overview. At client-side, each client mainly consists of two components: a local model $\theta$, a client-specific model $\phi$ to capture client information. At server-side, we adopt a client-centric selection strategy to sample the optimal client set for each client based on the client-specific representations extracted from client-specific model $\phi$ in the previous step.
 }
  \label{fig:framework} 
\end{figure*}

\section{PROBLEM STATEMENT}

\subsection{Preliminaries}
For conventional federated learning with $N$ clients, the objective is to learn a single model $w$ that minimize the loss function on $\mathcal{D}$ by optimizing the following expected risk:
$\min _w \mathcal{E}_{\mathcal{D}}(w) \approx \frac{1}{N}\sum_{i=1}^N  \mathcal{F}(\mathcal{D}_i;w),$ where error $\mathcal{F}(\mathcal{D}_i;w)$ is the local objective function for $i$-client. Let client data be represented by $\mathcal{X} \subset\mathbb{R}^D$, while $\mathcal{Y}$ denotes the label space. Data sample $(x,y)\sim \mathcal{D}: x \in \mathcal{X}, y \in \mathcal{Y}$, and $f_\mathcal{D}: \mathcal{X} \to \mathcal{Y}$ is a deterministic labeling function. In this paper, we consider a non-IID federated setting where local data distributions differ among clients.

\noindent\textbf{Domain Shift} \textit{For each client dataset $S_i$, the data distribution $\left(x_i, y_i\right) \sim \mathcal{D}_i$, consider a set of $k$ latent domains, each $S_i$ contains images from only one of the $G$ latent domains, such that $S_i \in\left\{1, \ldots, G_k\right\}$. The client datasets are not complete disjointed, such that there is domain-wise overlap across datasets: $\emptyset \neq \bigcap_{i=1}^I \mathcal{G}_{S_i}$, where $\mathcal{G}_{S_i}$ is the collection of all client datasets' latent domains.}

The $G$ latent visual domains differ in styles, rotation degrees and cities etc., in this case the model for one domain may no longer compatible to another domain. 

\noindent\textbf{Label Shift} \textit{For each client dataset $S_i$, the data distribution $\left(x_i, y_i\right) \sim \mathcal{D}_i$, consider a whole label set $\mathcal{Y}$ and a  $i_{t h}$ client label set $Y_i$, where $Y_i$ is a subset of $\mathcal{Y}$: $\{Y_1, \ldots,Y_n\}$. The client label set $Y_i$ varies across clients while not completely disjoint, such that there is class-wise overlap across client label sets: $\emptyset \neq \bigcap_{i=1}^N Y_i$.}

\subsection{Problem Formulation} \label{problemsetting}
Our goal is to obtain a personalized local model for each client. Hence, given a set of $N$ clients with $K_i$ data points $S_i =\left\{\left(x_i^j, y_i^j\right)\right\}_{j=1}^{K_i}$, assume there exists domain shift and label shift among client data distribution $\left(x_i, y_i\right) \sim \mathcal{D}_i$.  The overall objective function is to learn a model parameterized by $w$ that minimize the loss over each client distribution $\mathcal{D}_i$  by optimizing the following expected risk:
\begin{equation}
\min _w \mathcal{E}_{\mathcal{D}}(w) \approx \frac{1}{N} \sum_{i=1}^N \mathcal{F}_i(\mathcal{D}_i;w),
\end{equation}
where $\mathcal{F}_i(\mathcal{D}_i;w):=\mathbb{E}_{\left(x_i, y_i\right) \sim \mathcal{D}_i}\left[\ell\left(f\left(x_i\right), y_i\right)\right]$, $f(x_i)$ is the mapping function from the input to the predicted label, and $\ell$ is the loss function that penalizes the distance of $f(x_i)$ from $y_i$.

\section{METHODOLOGY}
In this section, following the problem setting stated in \textit{section} \ref{problemsetting}, we first introduce the personalization problem of the heterogeneous federated learning. Then we describe in detail our proposed algorithm by presenting the client selection strategy and the adaptation techniques as described in Alg.  \ref{alg:overall}. 

At client-side, each client mainly consists of two components: a local model $\theta$, a client-specific model $\phi$ to capture client information. During the training process local model $\theta$ on each client enhances knowledge transfer by learning simple representations. Simultaneously, the client-specific model $\phi$ maps samples to private feature spaces with characteristic embeddings. At server-side, we adopt a client-centric selection strategy to sample the optimal client set for each client based on the client-specific representations extracted from client-specific model $\phi$ in the previous step. Figure \ref{fig:framework} provides an overview of our proposed federated multi-source adaptation algorithm (FedCCA).

\subsection{Local and Client-specific Model}\label{sec:local training}
In our proposed method, we stress on the client-centric information, which requires local model and client-specific model to incorporate client-specific embeddings and domain-invariant representations at the same time.  We employ an additional model on each client for such purpose.
For client local training, let $C$ represents the number of classes within the client datasets. We here define the function of the $i_{t h}$ client local model as $f_{\theta}^{(i)}$, where $f_{\theta}^{(i)}: \mathbb{R}^X \rightarrow \mathbb{R}^C$.
For local updating, consider $n$ clients are selected on round $t$ to update their current local
model $\theta_i$ on private data $S_i =\left\{\left(\mathbf{x}_i^j, y_i^j\right)\right\}_{j=1}^{K_i}$ for fixed epochs. To obtain optimal classification performance on $i_{t h}$ client, the cross entropy loss function is defined as follows:
\begin{equation}
\label{localloss}
\mathcal{L}_{local}^{(i)} =  - a^{(i)}_t\sum_{j=1}^{K_i} y_j \log f_{\theta}^{(i)}(x_j),
\end{equation}
where $A = \{a_t^{(i)}\}_{i\in\mathcal N}$ is a binary integer variable controlling the selection status of client $i$. The same for the $i_{t h}$ client client-specific model $f_{\phi}^{(i)}$, where $f_{\phi}^{(i)}: \mathbb{R}^X \rightarrow \mathbb{R}^C$. However, during each communication round, the client-specific model is trained for fixed epochs regardless of whether the  $i_{t h}$ client is selected for participation. Such model does not participate in the aggregation process. The loss function of client-specific is defined as follows:
\begin{equation}
\label{eq:csloss}
\mathcal{L}_{CS}^{(i)} =  - \sum_{j=1}^{K_i} y_j \log f_{\phi}^{(i)}(x_j).
\end{equation}

\subsection{Client-centric Selection}

Particularly, a client-centric selection strategy is proposed to enhance knowledge transfer and aggregation shown in Alg. \ref{alg:selection}. As mentioned earlier, we believe that the client-specific information apart from domain-invariant information  is crucial. 
\begin{figure}[h] 
  \centering
  \includegraphics[width=0.9\linewidth]{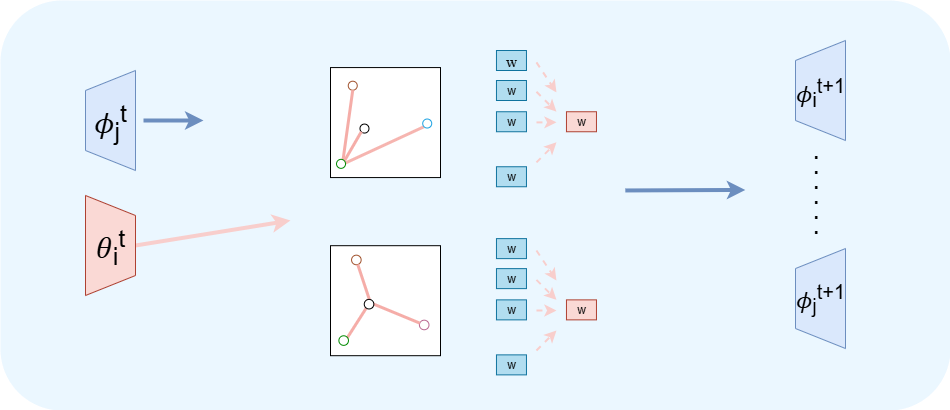}
  \caption{Illustration of client-centric adaptation in FedCCA.}
  \label{fig:adaptation} 
\end{figure} 
Accordingly client-centric adaptation integrate client information and collaboratively aggregate clients uploaded models. Yet we continue to refer to our approach as multi-source adaptation, for our objective of aligning various client data space shares similar fundamental principles with domain adaptation theory as previously stated. 
Owing to costly and 
\begin{algorithm}[htbp]

\label{alg:selection}
\SetKwData{Left}{left}\SetKwData{This}{this}\SetKwData{Up}{up} \SetKwFunction{Union}{Union}\SetKwFunction{FindCompress}{FindCompress} \SetKwInOut{Input}{Input}\SetKwInOut{Output}{Output}
\SetKwFunction{KwFn}{Fn}

\caption{The \textit{Client-Centric Selection} Algorithm}
\Input{Updated client-specific models $\{\phi_t^{(i)}\}$, Client set $\mathcal{N}$, Maximum selection $N^{max}$.}
\Output{Optimal client set $S^{(i)_t}$, Client status $\{a_{t+1}^{(i)}\}$, Attention-inducing function $\mathbf{u}_t^{(i,j)}$.}
\For{$t=0, 1, 2, ..., T$}{
Server received $\phi_t^{(i)}$ from all clients.\; 
\For{each client $i=1, 2, ..., N$ }{
\begin{flushleft}
$\mathbf{u}_t^{(i,j)}$ $\leftarrow$ Calculate function ($\phi_t^{(i)}, \phi_t^{(j)}$ ) by Eq. (5).\; 
$\mathcal{\text{Criteria}}_t^{(i)}$ $\leftarrow$ Calculate criteria ($\mathbf{u}_t^{(i,j)},N$ ) by Eq. (6).\; 
Construct the feasible client $S_t^{'(i)} = \subseteq \mathcal{N}$.
Rank the clients in $S_t^{'(i)}$ according to $\mathbf{u}_t^{(i,j)}$. 
Initialize $S_t^{(i)} = \emptyset$.\;
\While{$|S^{'(i)}| \leq N^{max}$}{
\If{$\mathbf{u}_t^{(i,j)} \leq \mathcal{\text{Criteria}}_t^{(i)}$}{
$S_t^{(i)} = S_t^{(i)} \cup \{n^{(j)}\}$.\;
$S_t^{'(i)}\backslash{n^{(j)}}$.
}
\Else{Break}
}
\end{flushleft}
}
}
\end{algorithm}
\noindent impractical communication with all devices in a large-scale federated system, a default strategy for client selection is sampling the clients uniformly at random and aggregating the model updates with weights proportional to the local samples \cite{MLSYS2020_38af8613}. As recent client selection researches highlight to eliminate inefficiency and redundancy of the federated system or solve heterogeneous problem through clustering, they employ sampling techniques from a systematic perspective. Unlike previous literature that cluster clients into fixed partitions and drop non-beneficial or less appropriate clients from sampling, we argue that client selection is not symmetric. Assume client $a,b \in S_c$ is in the optimal subset  of client $c$, denoted as $S_c$, while $a\notin S_b$ might possibly not in the optimal subset $S_b$ of client $b$. Our client-centric sampling strategy proposes to obtain an optimal client selection for each client, aiming at encouraging diverse adaptation at each communication round.

As shown in Fig. \ref{fig:adaptation}, the proposed client selection strategy is grounded in feature similarity comparison, where pairwise similarity is calculated to find a subset $S$ of clients over all the $N$ clientsat $t$ communication round.
We obtain the client-centric information through the fully-connected layer of the client-specific model $\phi_i$ through Eq.(\ref{eq:csloss}).
Taking simplicity and prevalence into consideration we adopt Euclidean Distance \cite{DANIELSSON1980227} as the metric.
Concretely, for the client-specific knowledge $\phi^{(i)}_{fc}$ of the $i_{t h}$ client and client-specific representation $\phi^{(j)}_{fc}$ of the $j_{t h}$ client, the Euclidean Distance can be measured by: 
\begin{equation}
\left\|\phi^{(i)}_{fc}-\phi^{(j)}_{fc}\right\|^2=\sqrt{\sum_{k=1}\left(E_i^k-E_j^k\right)^2},
\end{equation}
where $E_i^k$ and $E_j^k$ denote the $k_{t h}$ element in the parameters of $\phi^{(i)}_{fc}$ and $\phi^{(j)}_{fc}$ respectively. 
Extended upon \cite{Huang_Chu_Zhou_Wang_Liu_Pei_Zhang_2021}'s idea, we apply a term to evaluate the pairwise similarity between clients by an attention-inducing function $B\left(\left\|\phi^{(i)}_{fc}-\phi^{(j)}_{fc}\right\|^2\right)$ defined as follows. 
Given $B:[0, \infty) \rightarrow \mathbb{R}$ a non-linear function which is increasing and concave on $[0, \infty)$, continuously differentiable on $(0, \infty)$, and $B(0)=0$, 
The attention-inducing function $\mathbf{u}^{(i,j)}=B\left(\left\|\phi^{(i)}_{fc}-\phi^{(j)}_{fc}\right\|^2\right)$ measures the similarity between $\phi^{(i)}_{fc}$ and $\phi^{(j)}_{fc}$ in a non-linear manner.
We adopt the widely-used negative exponential function

\begin{equation}
B\left(\left\|\phi^{(i)}_{fc}-\phi^{(j)}_{fc}\right\|^2\right)=1-e^{-\left\|\phi^{(i)}_{fc}-\phi^{(j)}_{fc}\right\|^2 / \sigma},
\end{equation}
with a hyperparameter $\sigma$.
At each communication round $t$ we set a grouping criteria $\mathcal{\text{Criteria}}_i^{t}$ calculated by

\begin{equation}
\mathcal{\text{Criteria}}_t^{(i)}=\frac{1}{2N}\sum_{i\neq j}^{N} B\left(\left\|\phi^{(i)}_{fc}-\phi^{(j)}_{fc}\right\|^2\right).
\end{equation}

The server filters out the calculated pairwise similarity that is larger than the $\mathcal{\text{Criteria}}_i^{t}$ and a selected client set $S_i^t$ with more similar data distribution to the target $i_{t h}$ client at each communication round $t$ is picked out to further cooperatively update the client model $\theta^{(i)}_t$.

\subsection{Multi-source Aggregation Strategy}
For the client model updates, we propose an attentive multi-source adaptation strategy to enhance personalization. On the basis of the client-centric selection, at each communication round $t$, client $i$ identifies an optimal subset of $n$ clients, denoted as $S_i$, based on specific selection criteria. During the global aggregation phase, we employ a multi-source adaptation approach to update the local model of each client. Specifically, the updated model for client $i$ is computed as follows: 
\begin{equation}
\begin{split}
\theta_{t+1}^{(i)}= & \left(1-a_t^{(1)}B\left(\left\|\phi^{(i)}_{fc}-\phi^{(1)}_{fc}\right\|^2\right)\right) \cdot \theta_t^{(1)} +\cdots \\
& \cdots +\left(1-a_t^{(i)}B\left(\left\|\phi^{(i)}_{fc}-\phi^{(j)}_{fc}\right\|^2\right)\right) \cdot \theta_t^{(j)} \\
& = \mathbf{u}^{(1,i)}_{t}\theta_t^{(1)}+\cdots+\mathbf{u}^{(j,i)}_{t}\theta_t^{(j)},
\end{split}
\end{equation}
where $a_n^t\in\{0, 1\}, \forall t\in T, n\in  N$ is the selection status of source clients, $\mathbf{u}^{(j,i)}_{t}$ indicates the client similarity in the hidden space and presented as reference in the global aggregation for better adaptation. 
By leveraging knowledge from multiple similar clients, our approach enables each client to obtain a personalized federated model that is enriched with relevant information from its peers. This multi-source adaptation not only improves the generalization ability of the local models but also ensures that each client benefits from the collective knowledge of clients with similar data distributions, thereby achieving optimal personalization in federated learning. 

\begin{algorithm}[htbp]

\label{alg:overall}
\SetKwData{Left}{left}\SetKwData{This}{this}\SetKwData{Up}{up} \SetKwFunction{Union}{Union}\SetKwFunction{FindCompress}{FindCompress} \SetKwInOut{Input}{Input}\SetKwInOut{Output}{Output}
\SetKwFunction{KwFn}{Fn}

\caption{The \textit{FedCCA} Algorithm}\label{algorithm}
\Input{Initialized $\theta_0^{(i)},\phi_0^{(i)}, a^{(i)}_0$ for $N$ clients, Private data $(X_i,Y_i)$, Total communication round $T$, Local update epoch $E$, Learning rate $\eta$}
\Output{Optimal personalized model $\theta_t^{(i)}$}
\For{$t=0, 1, 2, ..., T$}{
Server sends $\theta_0^{(i)},\phi_0^{(i)}$ to all clients.\; 
\For{each client $i=1, 2, ..., N$ }{
\begin{flushleft}
$\theta_t^{(i)}$ $\leftarrow$ \textbf{\textit{Local Training ($\theta_{t}^{(i)}, a^{(i)}_t, E,\eta$ )}}.\;
$\phi_t^{(i)}$ $\leftarrow$ \textbf{\textit{Client-Specific Training ($\phi_{t}^{(i)},E,\eta$ )}}.\;
\end{flushleft}
}
$S^{(i)}_t,\{a_{t}^{(i)}\},\mathbf{u}_t^{(i,j)}$ $\leftarrow$ \textbf{\textit{Client-Centric Selection ($\{\phi_t^{(i)}\}$)}}. \; 
$\{\theta_{t+1}^{(i)}\}$ $\leftarrow$ \textbf{\textit{Multi-Source Aggregation ($\{\theta_t^{(i)}\}, \{\mathbf{u}_t^{(i,j)}\}$)}}.\;
}
 
\end{algorithm}

\section{EXPERIMENT}

\subsection{Datasets}
We perform experiments on five commonly used datasets in federated learning. We apply three of them for digit classification and two for object recognition to demonstrate the effectiveness of our proposed model. \textbf{Digits} is a collection of four benchmarks for digit recognition regarded as four different domains in our experiment, namely MNIST \cite{726791}, Synthetic Digits \cite{pmlr-v37-ganin15}, SVHN \cite{articleSVHN}, and USPS \cite{291440}. \textbf{RotatedMNIST} \cite{7410650} contains clockwise rotated MNIST images \cite{lecun-mnisthandwrittendigit-2010} at angles of
$0^{\circ}, 15^{\circ}, 30^{\circ}, 45^{\circ}, 60^{\circ}$ and $75^{\circ}$ forming six domains. \textbf{FEMNIST} \cite{caldas2019leaf} is built by partitioning the data in Extended
MNIST \cite{lecun-mnisthandwrittendigit-2010, cohen2017emnist} based on the writer of the digit or characters. \textbf{CIFAR-100} \cite{krizhevsky2009learning} is an image classification dataset with $100$ classes containing $600$ images each. The $100$ classes are grouped into $20$ superclasses. \textbf{DomainNet} \cite{Peng_2019_ICCV} is a large-scale dataset of common objects in six different domains including clipart, infograph, painting, quickdraw, real and sketch. The DomainNet comprehensively contains 345
classes and we use the top 100 most common classes to form a sub-dataset for our experiments.

\begin{table*}
\begin{center}    

\caption{The test accuracy (\%) on various partitions of Digit classification datasets}
\label{digit result}
\renewcommand{\arraystretch}{1.25}
\small
\setlength{\tabcolsep}{2.5mm}{
\begin{tabular}{lllcllcl}
\hline
METHODS                           & \multicolumn{2}{c}{DIGITS}           & \multicolumn{2}{c}{ROTATEDMNIST} & FEMNIST & \multicolumn{2}{c}{DOMAINNET}        \\ \cmidrule(r){2-3} \cmidrule(r){4-5} \cmidrule(r){6-6} \cmidrule(r){7-8}
\#CLASS/CLIENT & \multicolumn{1}{l}{($N=2$)} & ($N=5$) & \multicolumn{1}{l}{($N=2$)} & ($N=5$)& ($N=3$) & \multicolumn{1}{l}{($N=5$)} & ($N=20$) \\ \hline
FEDAVG \cite{Fedavg}               & 61.0$\pm$0.34 & 55.5$\pm$0.24 & 80.4$\pm$0.57 & 76.3$\pm$0.61 & 38.3$\pm$0.27 & 20.6$\pm$0.66                        & 23.4$\pm$0.47    \\
FEDPROX \cite{MLSYS2020_38af8613}               & 38.4$\pm$0.96                        &50.3$\pm$0.77    & 81.6$\pm$0.86                        & 77.4$\pm$0.45  & 19.5$\pm$0.45 & 20.1$\pm$0.63                        & 23.2$\pm$0.41    \\
FEDREP  \cite{collins2021exploiting}               & \textbf{79.5$\pm$0.34 }                       & 75.8$\pm$0.62    & 88.2$\pm$0.57                        & 83.9$\pm$0.69  & 64.6$\pm$0.03& 42.1$\pm$0.87                      & 37.5$\pm$0.46    \\
FEDAMP \cite{Huang_Chu_Zhou_Wang_Liu_Pei_Zhang_2021}                & 78.2$\pm$0.60                        & 75.1$\pm$0.47    & 87.3$\pm$0.09                        & 83.6$\pm$0.38  & 60.7$\pm$0.26 & 40.6$\pm$0.21                        & 34.1$\pm$0.79   \\ \hline
FEDDAN \cite{10.5555/3045118.3045130}                & 70.2$\pm$0.25                        & 64.2$\pm$0.76    & 85.4$\pm$0.64                        & 82.1$\pm$0.83  & 55.3$\pm$0.83& 36.9$\pm$0.46                        & 30.2$\pm$0.62     \\
FEDDANN(\cite{ganin2016domainadversarial}               & 72.8$\pm$0.74                       & 64.3$\pm$0.56    & 85.8$\pm$0.44                        & 82.3$\pm$0.61  & 56.9$\pm$0.53 & 37.4$\pm$0.74                        & 30.8$\pm$0.38   \\
FEDSR \cite{nguyen2022fedsr}                 & 79.3$\pm$0.79                       & 76.5$\pm$0.86    & 88.7$\pm$0.23                        & 83.8$\pm$0.47  & 65.1$\pm$0.21                    & \textbf{42.3$\pm$0.44}                        & 37.1$\pm$0.31    \\
FEDMC \cite{10.1145/3511808.3557378}                 & 78.0$\pm$0.51                       & 75.2$\pm$0.68    & 87.2$\pm$0.93                        & 83.3$\pm$0.47  &  64.8$\pm$0.78   & 41.5$\pm$0.49                        & 34.6$\pm$0.08    \\  \hline
FEDCCA (OURS)            & 78.2$\pm$0.24                       & \textbf{76.8$\pm$0.87}    & \textbf{88.9$\pm$0.13}                  & \textbf{83.9$\pm$0.80}  &\textbf{65.3$\pm$0.60} & 41.7$\pm$0.68                        & \textbf{38.3$\pm$0.32 }   \\ \hline
\end{tabular}}
\end{center}
\end{table*}

\begin{table}
\begin{center}    

  \caption{The test accuracy (\%) on various partitions of object recognition datasets}
  \label{image result}
\renewcommand{\arraystretch}{1.25}
\small
\setlength{\tabcolsep}{2.5mm}{
\begin{tabular}{lcllcl}
\hline
METHODS                & \multicolumn{3}{c}{CIFAR-100}        \\ \cmidrule(r){2-4} 
\#CLASS/CLIENT & \multicolumn{1}{l}{($N=5$)} & ($N=10$)   & ($N=20$)  \\ \hline
FEDAVG \cite{Fedavg}              & 22.7$\pm$0.74                    & 25.1$\pm$0.23 & 29.6$\pm$0.51 \\
FEDPROX \cite{MLSYS2020_38af8613}               & 20.4$\pm$0.35                    & 25.5$\pm$0.90 & 28.7$\pm$0.82 \\
FEDREP  \cite{collins2021exploiting}               & 78.4$\pm$0.61                    & 67.6$\pm$0.67 & 52.3$\pm$0.36 \\
FEDAMP \cite{Huang_Chu_Zhou_Wang_Liu_Pei_Zhang_2021}                      & 77.6$\pm$0.65                    & 64.5$\pm$0.36 & 51.6$\pm$0.98 \\ \hline
FEDDAN \cite{10.5555/3045118.3045130}                & 71.8$\pm$0.35                    & 62.4$\pm$0.64 & 45.9$\pm$0.01  \\
FEDDANN \cite{ganin2016domainadversarial}                  & 71.5$\pm$0.57                    & 62.3$\pm$0.86 & 45.3$\pm$0.69 \\
FEDSR \cite{nguyen2022fedsr}                         & 78.7$\pm$0.71                    & 67.7$\pm$0.65 & 52.5$\pm$0.23  \\
FEDMC \cite{10.1145/3511808.3557378}                   & 77.1$\pm$0.46                    & 64.0$\pm$0.27 & 51.6$\pm$0.24  \\ \hline
FEDCCA (OURS)             & \textbf{78.4$\pm$0.87}                   & \textbf{67.9$\pm$0.85} & \textbf{53.1$\pm$0.43} \\ \hline
\end{tabular}}
\end{center}
\end{table}

\subsection{Baselines}

We consider the following baselines to evaluate the model performance.
\textbf{FedAvg} \cite{Fedavg} is a foundational approach that enables clients to collaboratively train a global model by averaging their local parameters. \textbf{FedProx} \cite{MLSYS2020_38af8613} and \textbf{FedRep} \cite{collins2021exploiting} extends FedAvg by introducing a proximal term to regularize local updates and split the model training. \textbf{FedAMP} \cite{Huang_Chu_Zhou_Wang_Liu_Pei_Zhang_2021} employs an adaptive message passing algorithm to facilitate efficient communication and aggregation of client updates. We also compare methods that use domain adaptation techniques, \textbf{FedDAN} \cite{pmlr-v37-long15} aligns source and target domains using a multi-kernel Maximum Mean Discrepancy (MMD) loss and \textbf{FedDANN} \cite{pmlr-v37-ganin15} further reduces domain gaps by incorporating a gradient reversal layer. \textbf{FedSR} \cite{nguyen2022fedsr} is a representation learning framework designed for federated domain generalization. Lastly, \textbf{FedMC} \cite{10.1145/3511808.3557378}leverages a meta-critic to capture robust, domain-invariant knowledge across clients. These baselines collectively represent a diverse set of strategies for collaborative, personalized, and domain-adaptive federated learning.

\subsection{Experiment Setup}

To demonstrate the validity of our theoretical results, we perform experiments on five commonly used datasets in generalized federated learning.
For the three digit classification datasets, we use a similar yet slightly different CNN architecture as \cite{li2021fedbn}. A single
linear layer is then used to map the concatenated two representations to the output classes.  
For the CIFAR-100 dataset, we adopt the commonly used backbone network ResNet18 \cite{7780459} as the encoder networks and a ResNet50 \cite{7780459} backbone for
DomainNet. Following the design in \cite{nguyen2022fedsr}, we replace the last fully connected layer of the backbone with a linear layer form the representation network and another layer to map the input to the finite class. 
We use the cross-entropy loss and stochastic gradient descent (SGD) optimizer with learning rate 0.01 and 0.001 for object recognition and digit classification experiments respectively.
We set the communication round $T=100$ and for each local update, we execute $E=5$ local epochs with a batch size set to $32$.




\begin{figure}[t]
\setlength{\abovecaptionskip}{-0.05cm}
\centering
\begin{subfigure}
[FedCCA.]
{\includegraphics[width =0.23\textwidth]{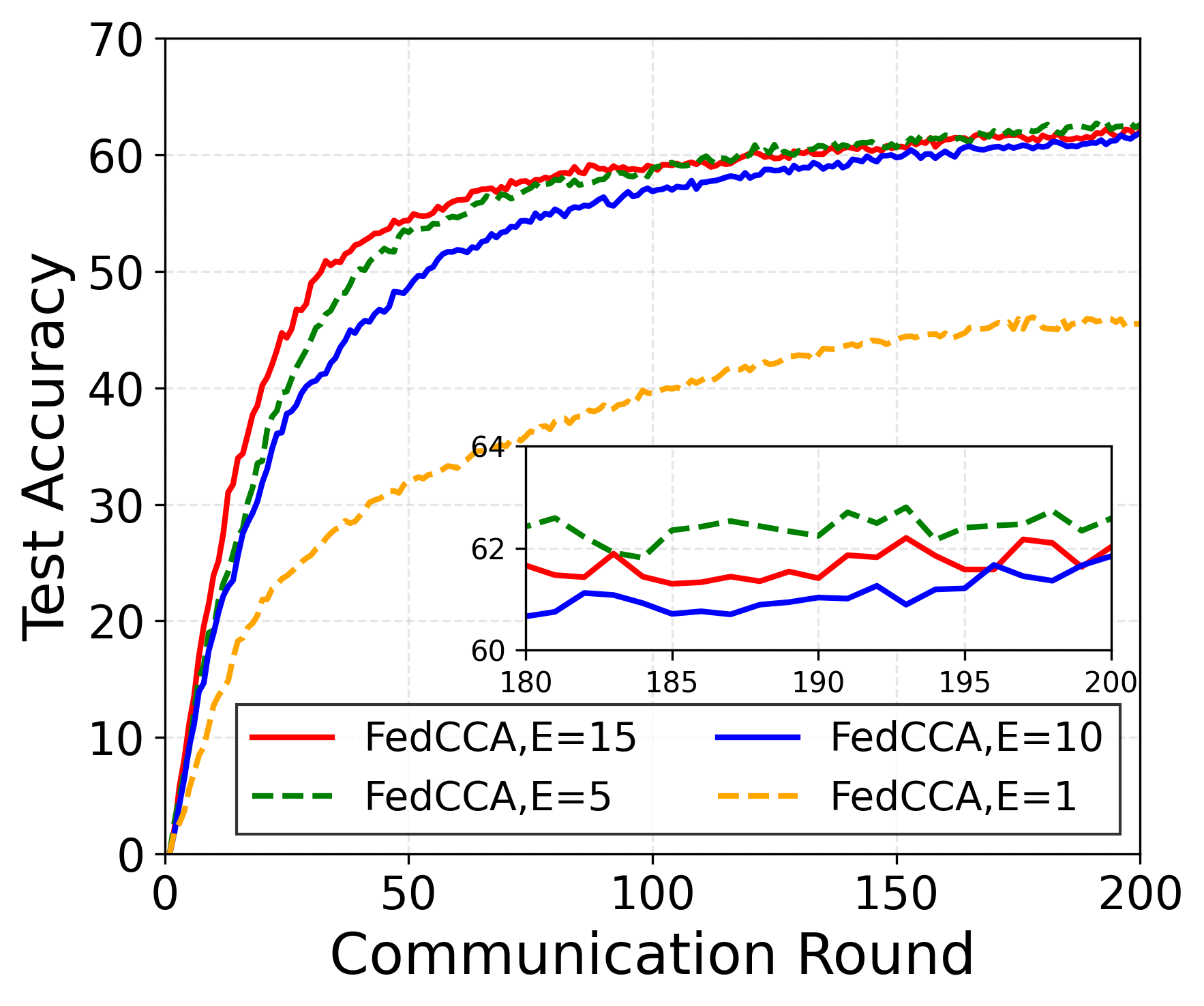}
\label{fig:epoca_cca}
}
\end{subfigure}
\hfill
\begin{subfigure}
[FedCCA and FedAVG]{\includegraphics[width=0.23\textwidth]{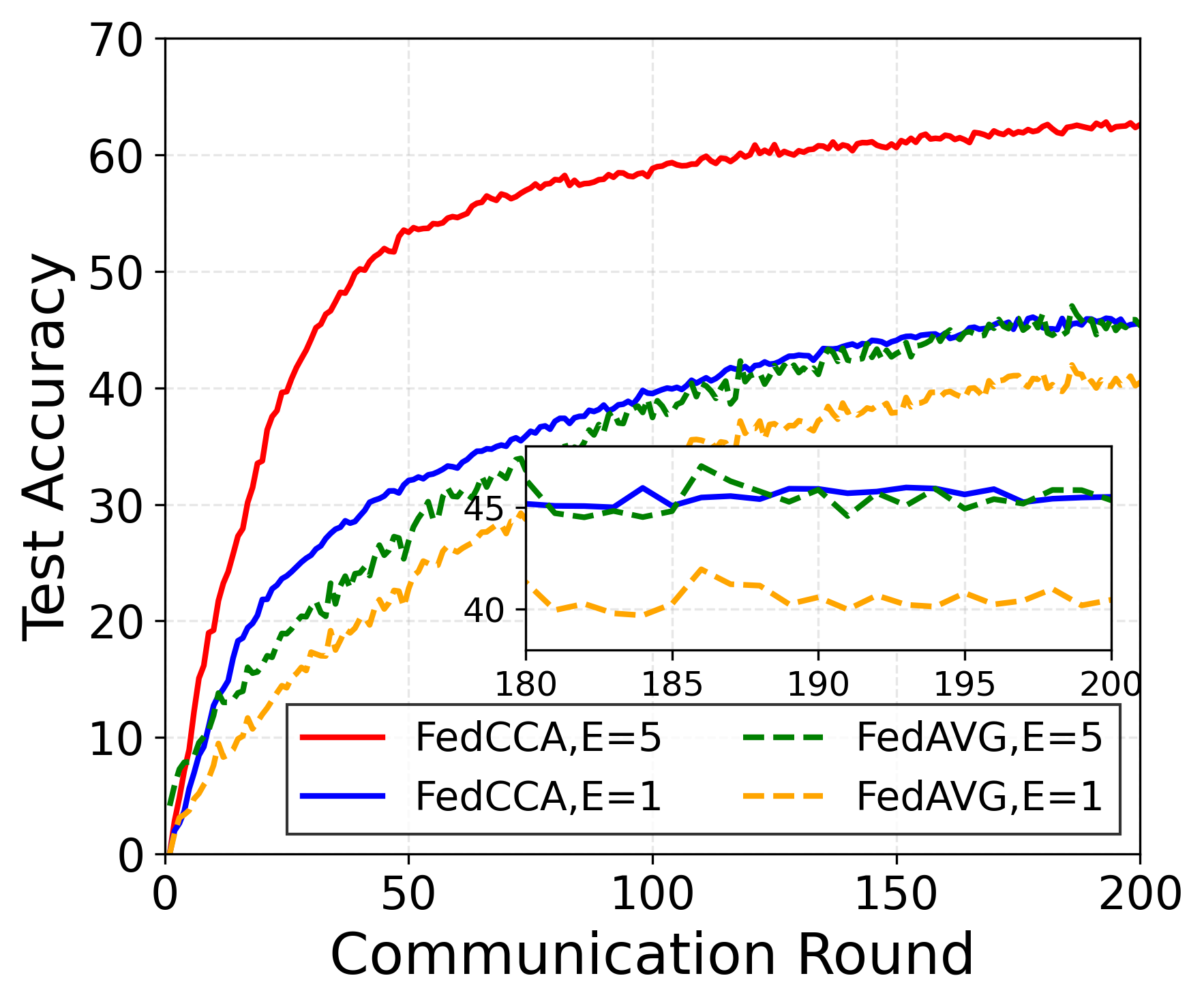}
\label{fig:epoch_2}
}
\end{subfigure}
\caption{Training performance over CIFAR-100 under different epochs.}
\label{fig:epoch}
\vspace{-0.1in}
\end{figure}

\begin{figure}[t]
\setlength{\abovecaptionskip}{-0.05cm}
\centering
\begin{subfigure}
[CIFAR-100.]
{\includegraphics[width =0.23\textwidth]{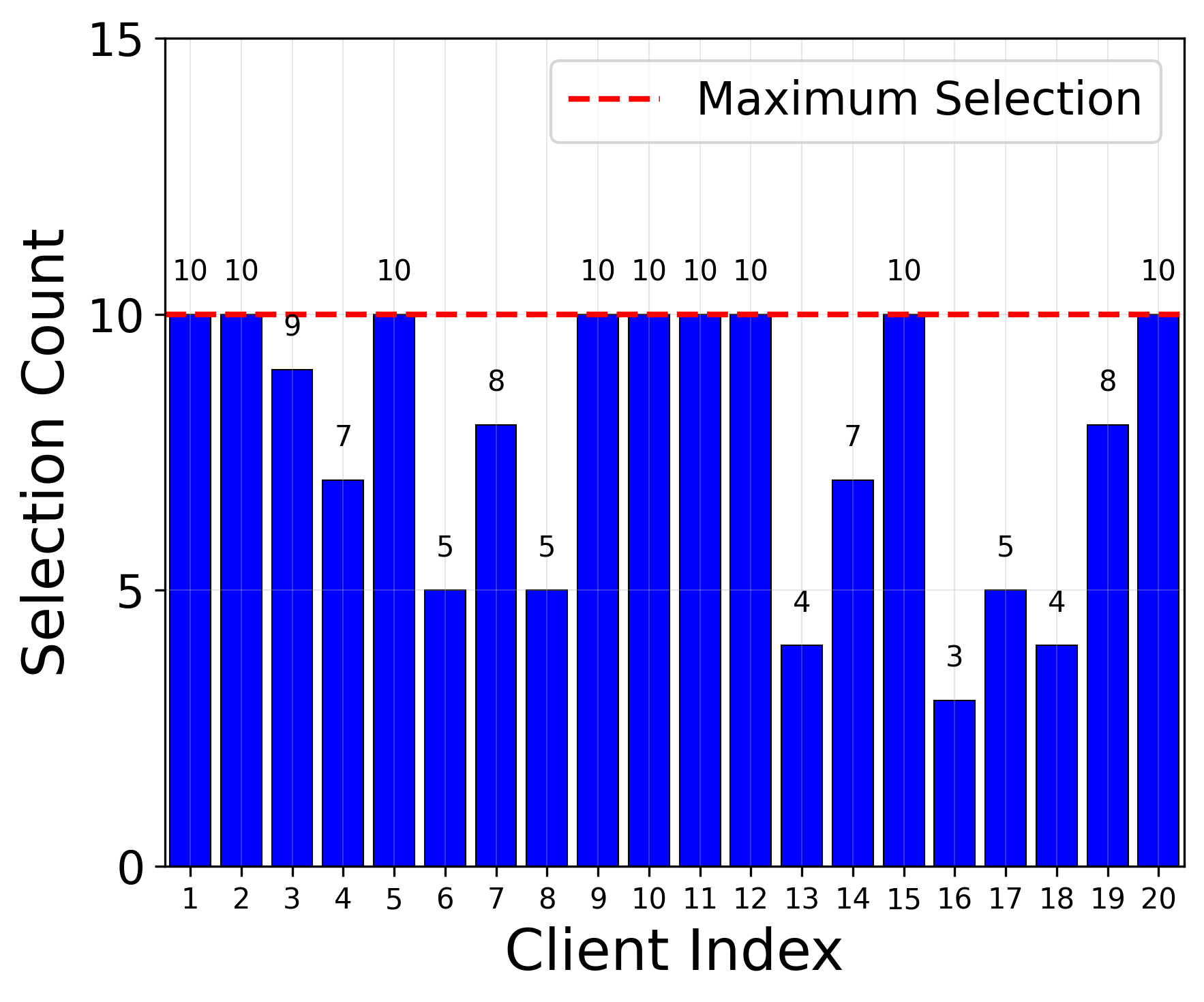}
\label{fig:selection_n}
}
\end{subfigure}
\hfill
\begin{subfigure}
[DomainNet.]{\includegraphics[width=0.23\textwidth]{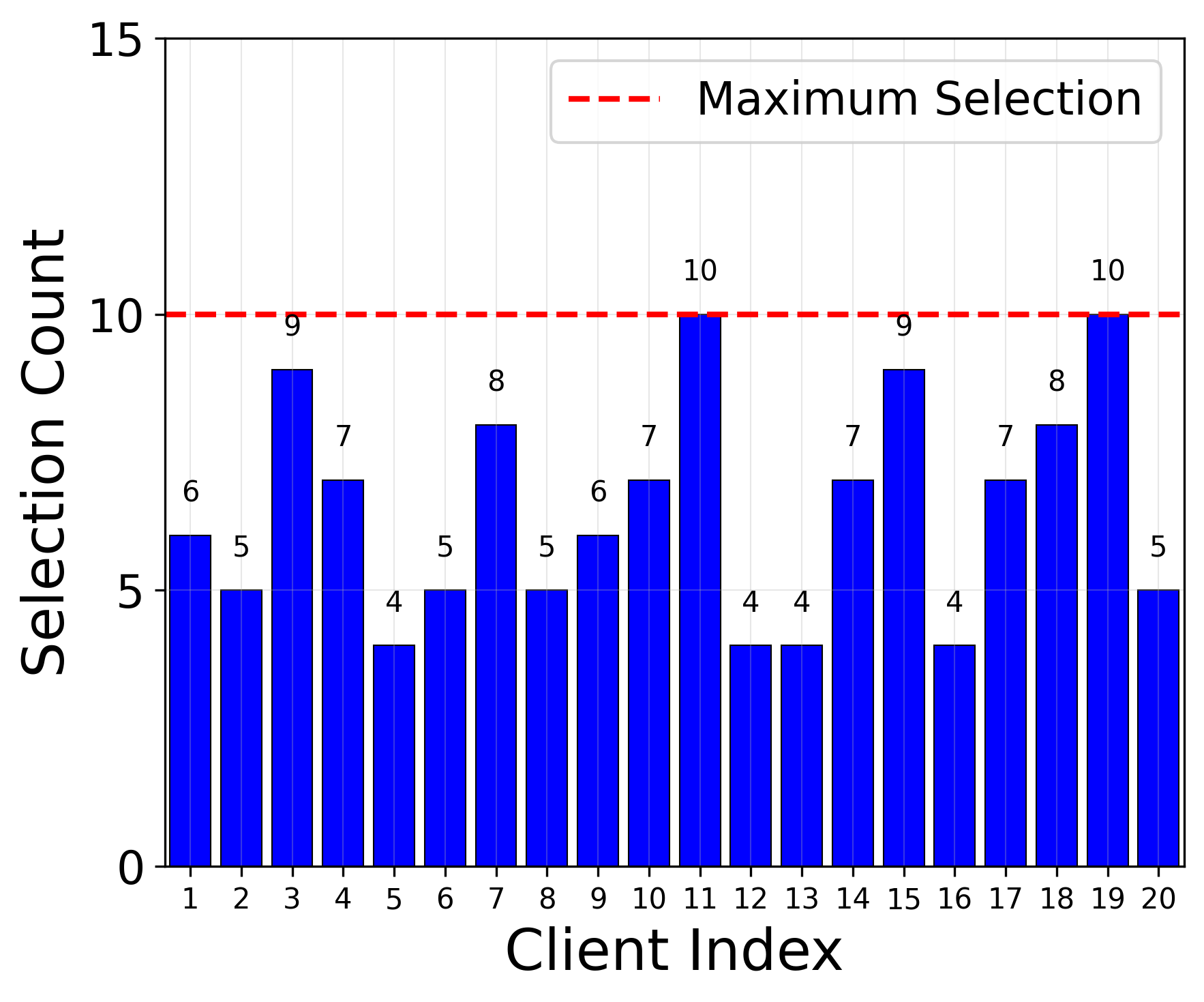}
\label{fig:selection_h}
}
\end{subfigure}
\caption{Client selection counts over different datasets.}
\label{fig:selection}
\vspace{-0.1in}
\end{figure}

\subsection{Evaluation of FedCCA}
We present comparative results on benchmark federated learning experiments on two classification tasks: digit classification and object recognition, using five publicly available datasets.
The experimental results of digit recognition are shown in Table \ref{digit result} and Table \ref{image result}. The results demonstrate that our method (FedCCA) surpassed the baselines on Digits, achieving a mean accuracy improvement compared to all other methods. 
The superior results of our method compared with other baseline methods employing domain adaptation to federated setting demonstrate the effectiveness of our method. 
We also find that non-personalized baseline methods such as FedAVG and FedProx perform much worse than other algorithms which train unique model for each client. Such discovery demonstrates the indispensable need for personalization in heterogeneous federated learning framework design.
Fig. \ref{fig:selection} presents an average client selection count over CIFAR-100 dataset as shown in Fig. \ref{fig:selection_n}, and a more heterogeneous dataset DomainNet in Fig. \ref{fig:selection_h}. The results demonstrate the effectiveness of FedCCA in adaptively selecting the optimal client subsets.

\begin{figure}[t]
\setlength{\abovecaptionskip}{-0.05cm}
\centering
\begin{subfigure}
[Heterogeneous case.]
{\includegraphics[width =0.23\textwidth]{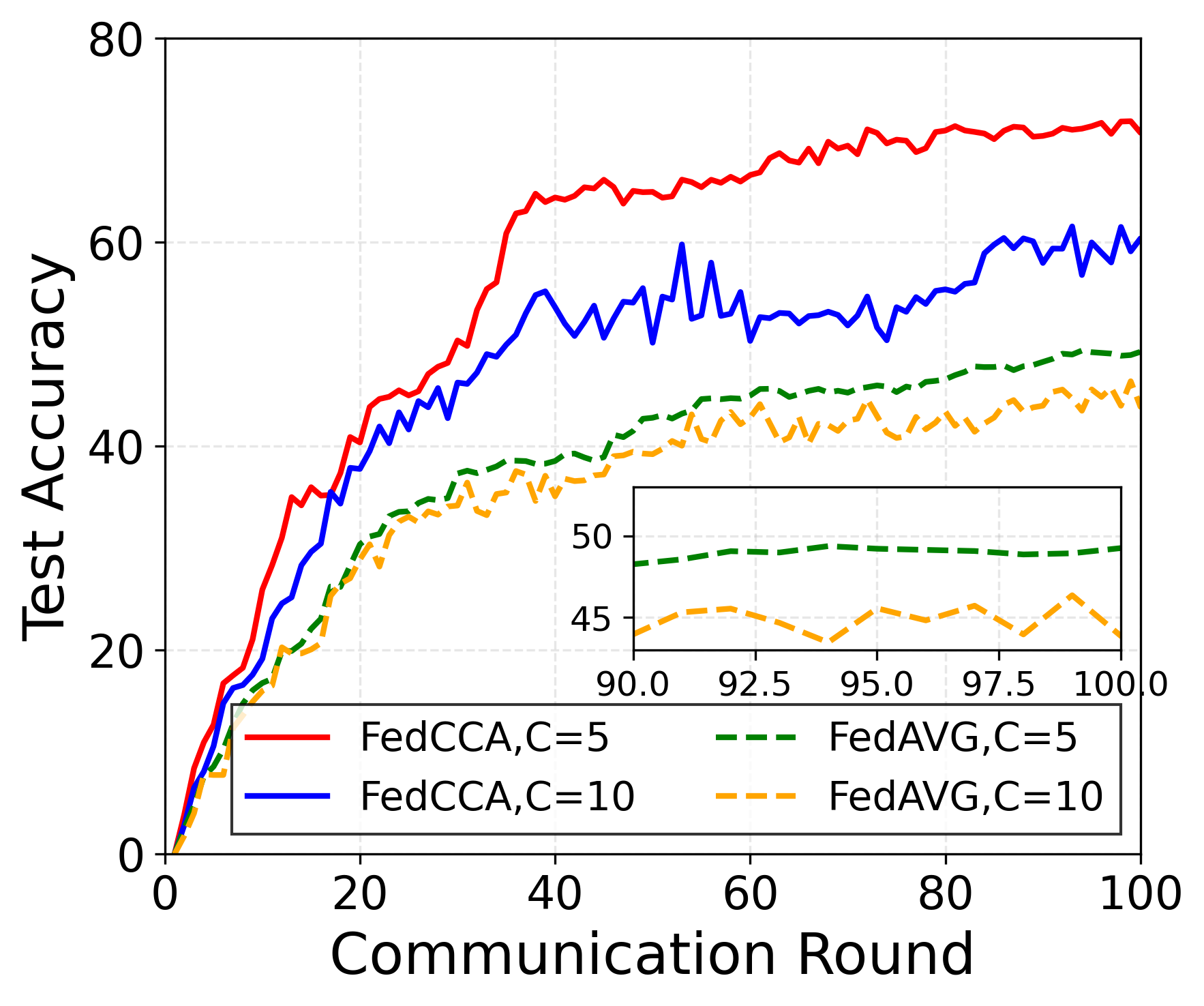}
\label{fig:hetero_n}
}
\end{subfigure}
\hfill
\begin{subfigure}
[Extreme case.]{\includegraphics[width=0.23\textwidth]{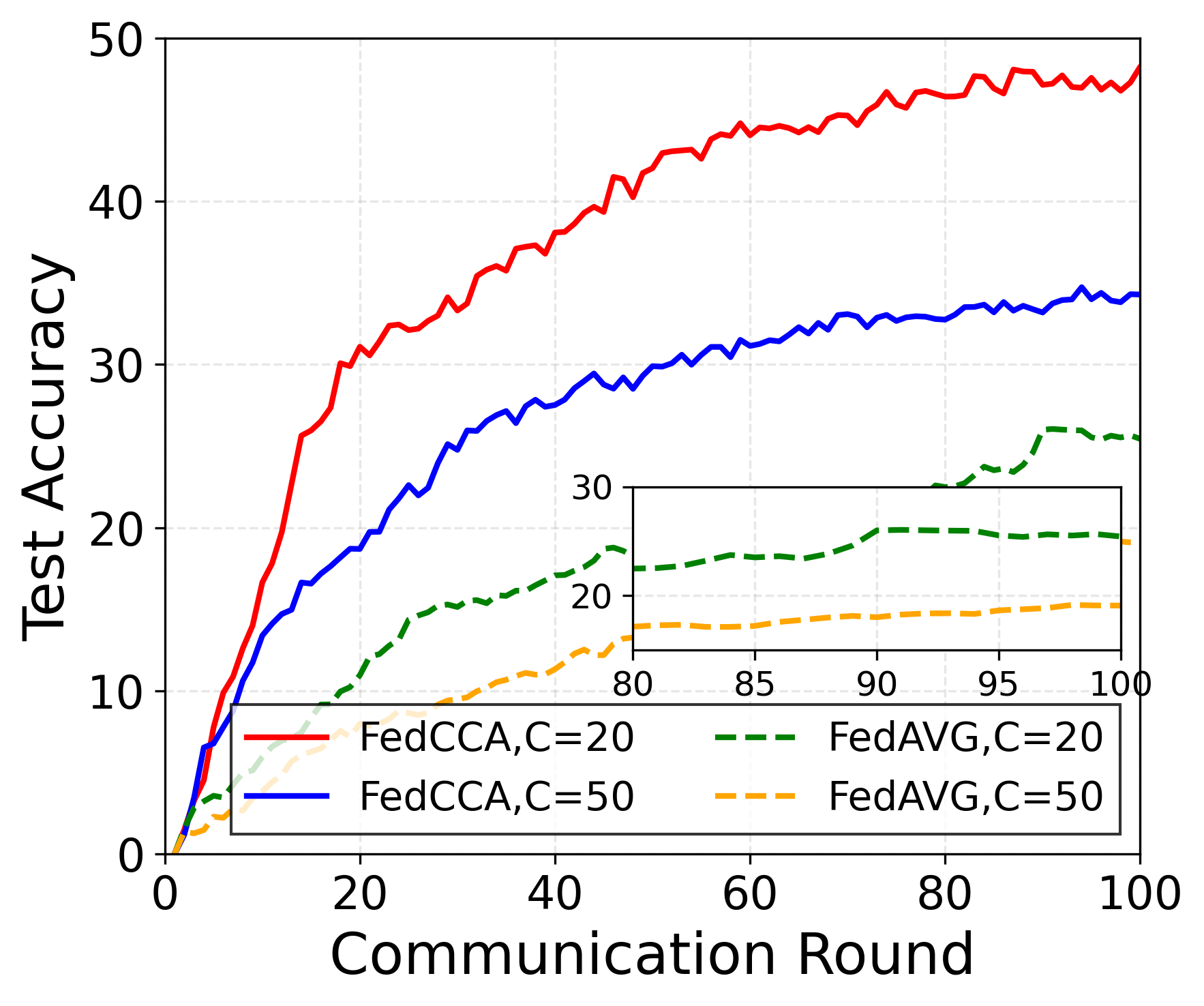}
\label{fig:hetero_h}
}
\end{subfigure}
\caption{Training performance over CIFAR-100 under different data heterogeneity.}
\label{fig:hetero}
\vspace{-0.1in}
\end{figure}

\subsection{Hyper Parameters Analysis}
\subsubsection{Effects of Local Updating} 

We explore the effects of local update epochs by comparing the accuracy of different baseline methods when the epoch is set from $\{1,5,10,15\}$ with fixed number of global communication rounds of $200$. This experiment involves simulating a heterogeneous setting using the CIFAR-100 dataset and $20$ clients. We maintain the rest of the experimental setup the same as previously stated. 
A relationship between the local updating epochs and testing accuracy implied for both FedCCA is visualized in Fig. \ref{fig:epoca_cca}. We notice that increasing the number of local updates not only leads to improved performance but also imperatively benefit the convergence. A comparison between FedCCA and baseline method is shown in Fig. \ref{fig:epoch_2}. Upon analysis, we observed that FedCCA consistently outperformed FedAVG by a significant margin.

\subsubsection{Effects of Data Heterogeneity} 

To study the effects of data heterogeneity, we conduct experiments across varying levels of statistical heterogeneity. The results show that our method FedCCA strongly outperforms others in practical non-IID settings.
We control the client data generation process with different concentration parameters $\alpha$ chosen from $\{0.2,0.4,0.6,0.8\}$ of the Dirichlet distribution as shown in Fig. \ref{fig:hetero_alpha}.
The results show that FedCCA obtains a significant improvement in comparison with benchmark algorithm in most cases. 
We also evaluate FedCCA under pathological non-IID settings where each client contains data from limited classes controlled by hyper-parameter $N$ in Fig. \ref{fig:hetero}.
Fig. \ref{fig:hetero_n} illustrate a heterogeneous case while Fig. \ref{fig:hetero_h} presents an extreme case of non-IID client data. The result shows that FedCCA shows a superior ability in dealing with severe heterogeneous settings, demonstrating the effectiveness of the proposed multi-source adaptation technique.

\subsection{Ablation Study}

\begin{figure}[t]
\setlength{\abovecaptionskip}{-0.05cm}
\centering
\begin{subfigure}
[Heterogeneous case.]
{\includegraphics[width =0.23\textwidth]{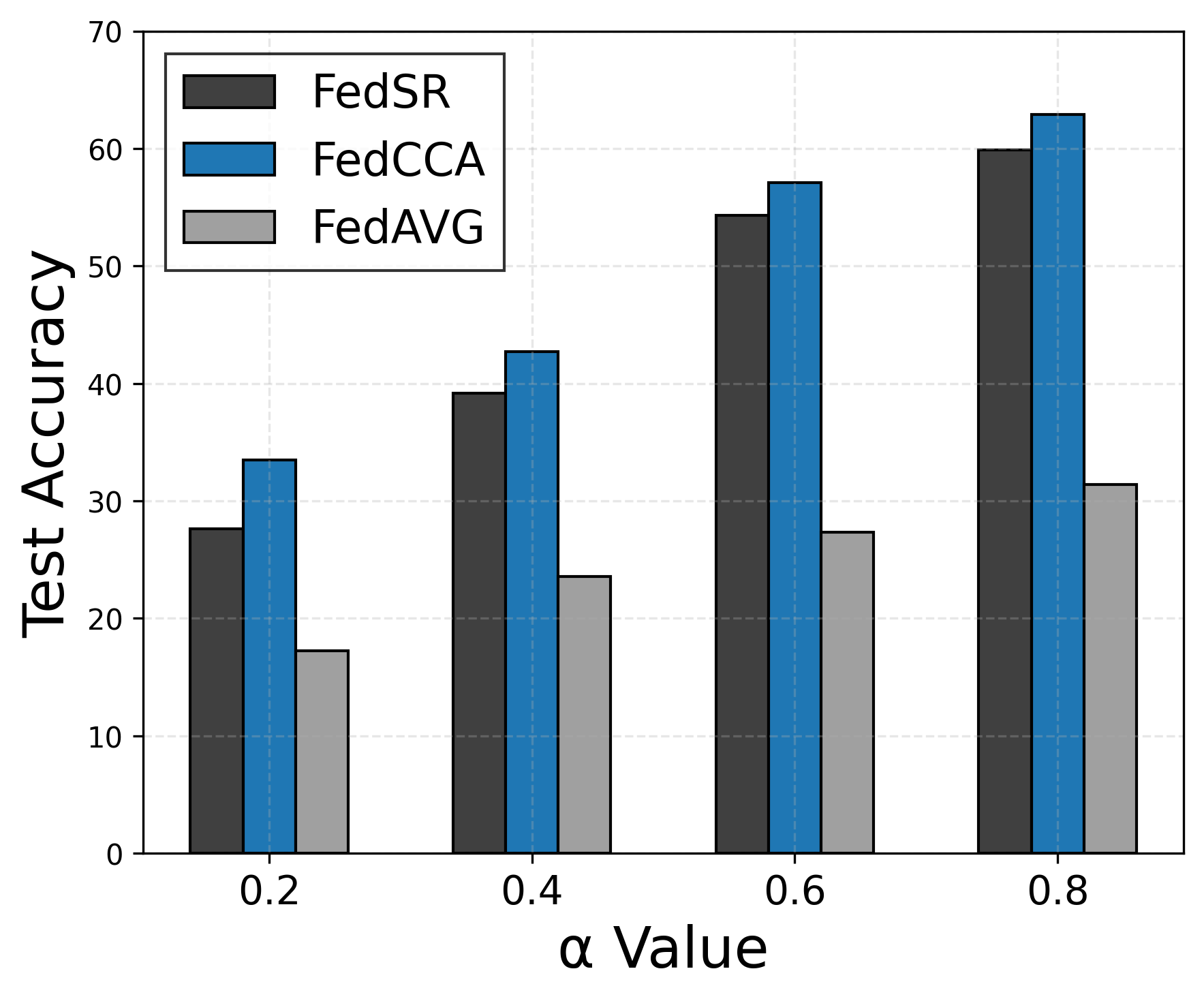}
\label{fig:hetero_alpha}
}
\end{subfigure}
\hfill
\begin{subfigure}
[Ablation studies.]{\includegraphics[width=0.23\textwidth]{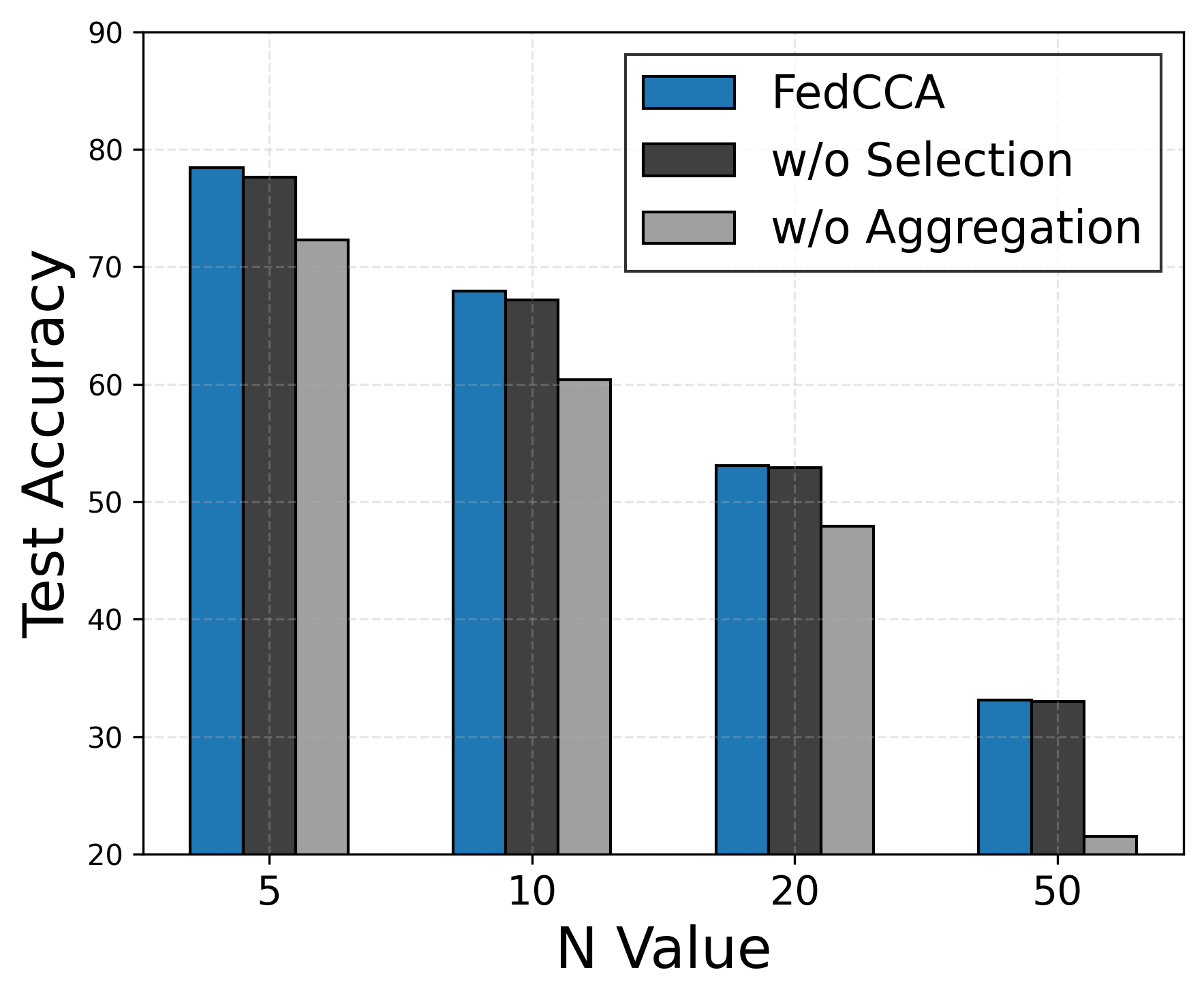}
\label{fig:ablation}
}
\end{subfigure}
\caption{Training performance over CIFAR-100 under different settings.}
\label{fig:further}
\vspace{-0.1in}
\end{figure}

In the previous discussion, our method provides superior performance in various image classification tasks. To gain deeper insights into the client-centric selection and multi-source aggregation utilized in our model, we conducted experiments on CIFAR-100 dataset by running our algorithm with different combinations.
As shown in Fig. \ref{fig:ablation}, 
When the multi-source aggregation module is omitted, a significant decline in performance is observed, indicating its critical role in the overall system. In contrast, the inclusion of the client selection module exerts only a marginal effect on overall accuracy. However, it substantially increases the operation time of the global server due to the additional computational requirements introduced by the selection process.
The study demonstrates that the proposed method FedCCA yields the most substantial performance improvement, highlighting the importance of client-centric federated learning.

\section{CONCLUSION AND DISCUSSION}
In this work, we addressed the critical challenge of data heterogeneity in federated learning within the Internet of Things (IoT) environments, where non-IID data distributions and privacy concerns pose significant obstacles to effective model training. We introduced FedCCA, a novel client-centric adaptation framework that leverages dynamic client selection and attention-based global aggregation to enhance personalized model performance.
Our extensive experiments on real-world datasets demonstrate that FedCCA consistently outperforms existing baseline methods, achieving superior local model accuracy and convergence rates across diverse tasks. These results highlight the importance of client-centric strategies in federated learning, particularly in heterogeneous and privacy-sensitive IoT scenarios. Furthermore, our findings underscore the necessity of moving beyond static, system-level optimization approaches to embrace dynamic, personalized aggregation and selection techniques.
Overall, FedCCA provides a robust and practical solution for federated learning in real-world IoT applications, bridging the gap between privacy preservation and model personalization. Our future work will extend our framework to broader IoT and edge computing contexts, further advancing the applicability and scalability of federated learning in heterogeneous environments.

\bibliography{reference}
\bibliographystyle{IEEEtran}

\end{document}